\documentclass[runningheads]{llncs}
\usepackage{siunitx}
\usepackage{booktabs}
\usepackage{subcaption} 
\usepackage[T1]{fontenc}
\usepackage{amsmath}
\usepackage{amssymb}
\usepackage{graphicx,verbatim}
\usepackage[ruled,vlined]{algorithm2e}
\usepackage{amsmath}
\usepackage{graphicx,verbatim}
\usepackage{graphicx}
\usepackage{color}
\usepackage{hyperref}
\begin{document}
\title{Depth-Augmented and FE-Free 3D--2D Liver Registration for Laparoscopic Liver AR}
%
\author{
Hanyuan Zhang\inst{1} \and
Lucas He\inst{1,3} \and
Runlong He\inst{1} \and
Weixi Yi\inst{1} \and
Abdolrahim Kadkhodamohammadi\inst{4} \and
Danail Stoyanov\inst{1} \and
Brian R. Davidson\inst{1,2} \and
Evangelos B. Mazomenos\inst{1} \and
Matthew J. Clarkson\inst{1}
}

\authorrunning{H. Zhang et al.}

\institute{
UCL Hawkes Institute, University College London, London WC1E 6BT, UK \\
\email{hanyuan.zhang.23@ucl.ac.uk}
\and
Division of Surgery and Interventional Science, University College London, London WC1E 6BT, UK
\and
Unit for Lifelong Health and Ageing at UCL, University College London, London WC1E 7HB, UK
\and
Medtronic plc., London, UK
}
  
\maketitle              
\begin{abstract}
Augmented reality (AR) guidance in laparoscopic liver surgery requires accurate registration of preoperative 3D models to intraoperative 2D video, but remains challenging due to partial visibility, specularities, and tissue deformation. Existing methods often rely on contour-based rigid initialization and finite-element (FE) models for deformable registration, increasing modeling and engineering complexity.
We present a depth-augmented, FE-free 3D--2D registration pipeline that combines robust rigid initialization with patient-specific non-rigid refinement. For rigid alignment, we adapt the RefineNet module of FoundationPose to laparoscopic liver scenes by using multi-class contour maps and monocular depth for relative pose refinement. For deformable alignment, we construct a patient-specific statistical deformation model from non-rigid ICP (NICP) correspondences and optimize pose and shape parameters using a coarse-to-fine L-BFGS-B strategy.
On a public clinical laparoscopic liver dataset, the proposed method achieves a mean target registration error (TRE) of 14.73\,mm under a controlled manual-contour setting designed to isolate registration performance. Ablation studies show that monocular depth improves rigid initialization over contour-only inputs, while tumor-mapping analysis indicates that good surface alignment does not necessarily translate into lower target localization error. On an external dataset without ground truth, the method produces visually plausible overlays for qualitative assessment. These results suggest that depth-augmented pose refinement and FE-free statistical deformation modeling provide a promising alternative to FE-based pipelines for controlled 3D--2D liver registration in surgical AR.
\end{abstract}

%
%
%
\section{Introduction}

In laparoscopic liver surgery, localizing tumors and major vessels is difficult because critical structures are not directly visible on the organ surface. Intraoperative ultrasound can partially mitigate this limitation, but it is operator-dependent and increases cognitive load. Augmented reality (AR) guidance addresses this challenge by overlaying preoperative computed tomography (CT) onto endoscopic views to visualize critical anatomy in situ, improving surgical safety and precision \cite{ramalhinho2023,schneider2020comparison}.Accurate overlays remain challenging because the liver deforms intraoperatively and only a limited surface region is visible, causing registration errors to propagate to tumor and vessel localization.\cite{ali2025objective}.

Registering preoperative 3D liver models to 2D intraoperative images typically begins with rigid initialization, commonly aligning projected 3D contours to 2D segmentations \cite{koo2022automatic}. While intraoperative depth is frequently used to reconstruct partial surfaces for explicit 3D-3D registration\cite{labrunie2023automatic}, this intermediate step is computationally heavy and fragile under sparse visibility. Bypassing 3D reconstruction to directly fuse monocular depth as a 2D geometric cue for \emph{3D-2D initialization} remains underexplored \cite{huang2025landmark}. Recent learning pipelines offer fast initialization but rely heavily on contours \cite{zhang2025deep}, which suffer from view-direction ambiguities under occlusion or deformation. As monocular depth estimation becomes increasingly reliable \cite{pei2024depth,cui2026depth}, its relative geometry can effectively disambiguate coarse 3D orientation (e.g., which lobe is closer) without the overhead of point-cloud generation. This depth-aware initialization provides a robust starting point for subsequent non-rigid refinement, preventing poor local minima.

The second stage is non-rigid registration. Classical methods such as non-rigid ICP (NICP) and Coherent Point Drift (CPD)~\cite{kundu2024comparative} often require intraoperative 3D surface reconstruction, which is difficult to obtain robustly under partial visibility. Physics-based finite-element (FE) pipelines~\cite{labrunie2023automatic} can model complex deformations, but they increase computational and engineering complexity and depend on uncertain patient-specific mechanical parameters. In liver surgery, tissue stiffness can vary substantially across patients and disease conditions, including cirrhosis, steatosis, and chemotherapy-induced fibrosis~\cite{lemine2024mechanical}. These limitations motivate computationally efficient and anatomically adaptive alternatives that do not rely on FE modeling.

Inspired by FoundationPose~\cite{wen2024foundationpose}, which leverages RGB-D cues for robust pose tracking, we adapt this idea to laparoscopic liver registration by combining multi-class contour cues with monocular depth for rigid pose initialization. Building on this initialization, we introduce an FE-free patient-specific deformation model for non-rigid registration under partial visibility. We further analyze how surface deformation transfer affects downstream tumor localization, showing that accurate surface fitting alone is insufficient for clinically meaningful AR guidance. Our main contributions are:
\begin{enumerate}
\item A depth-augmented rigid initialization method that adapts FoundationPose RefineNet to laparoscopic liver registration by fusing multi-class contour cues with monocular depth, improving robustness under partial visibility.
\item An FE-free non-rigid registration framework based on a patient-specific statistical deformation model learned from NICP-derived deformation correspondences, enabling low-dimensional shape refinement without explicit biomechanical simulation.
\item An analysis of tumor localization after non-rigid registration, showing that improved surface alignment does not necessarily translate into lower target registration error, thereby highlighting deformation transfer as a critical factor for clinically relevant AR guidance.
\end{enumerate}
\section{Method}
\noindent\textbf{Depth-Augmented Pose Initialization:}
We adapt the RefineNet module of FoundationPose (Wen et al.~\cite{wen2024foundationpose}) to predict a relative pose offset for rigid initialization in laparoscopic liver scenes. 
Specifically, we modify only the input interface by replacing the first convolution to accept a 5-channel tensor (4-class contour maps + masked monocular depth), while keeping the remaining architecture unchanged. 
The network is trained from scratch on synthetic renderings. 
Given an observed frame, the network regresses a 6-DoF pose update that aligns a rendered liver hypothesis to the observation. 
To reduce appearance-induced ambiguity and the synthetic-to-real gap, we replace RGB with geometry-aware cues and apply modality-specific augmentations (contour thinning/dilation with dropout/occlusion; depth erasing and scale perturbation), using:
\begin{enumerate}
    \item \textbf{Multi-class contour maps}, including right ridge, left ridge, ligaments, and silhouette;
    \item \textbf{Depth map} of the liver surface with instrument-occluded regions removed, estimated by Depth Anything V2 (Yang et al.~\cite{yang2024depth}).
\end{enumerate}

Depth map (D) denotes \emph{masked} monocular depth (depth \(\times\) liver mask). Instead of supervising rotation and translation independently using MSE as in the original implementation, we adopt a \emph{surface MSE} that directly penalizes misalignment of the transformed 3D liver surface. Let $\mathbf{T}_A$ and $\mathbf{T}_B$ denote the camera-to-liver transformations associated with the rendered hypothesis (input $A$) and the target intraoperative observation (input $B$), respectively. The network predicts a \emph{relative} transform $\mathbf{T}_{\mathrm{pred}}$ such that $\mathbf{T}_B \approx \mathbf{T}_A \mathbf{T}_{\mathrm{pred}}$. During inference, the network estimates this relative pose offset in a single forward pass (one-shot), avoiding the computational overhead of iterative refinement. For mesh vertices $\{\mathbf{p}_i\}_{i=1}^{N}$ (homogeneous coordinates), we define
\begin{equation}
\mathcal{L}_{\mathrm{surface}}
= \frac{1}{N} \sum_{i=1}^{N}
\left\|
\mathbf{T}_A \mathbf{T}_{\mathrm{pred}} \mathbf{p}_i
- \mathbf{T}_B \mathbf{p}_i
\right\|_2^2,
\end{equation}
where $N$ can be the full vertex set or a uniform subsample for efficiency. In our benchmark experiments, contour maps are obtained from manual annotations to isolate registration performance; in a fully automatic pipeline, they can be produced by existing segmentation/landmark networks~\cite{pei2024depth,xu2025lapfm,lin2025subsampled}.

\begin{figure}[t]
    \includegraphics[width=1.0\linewidth]{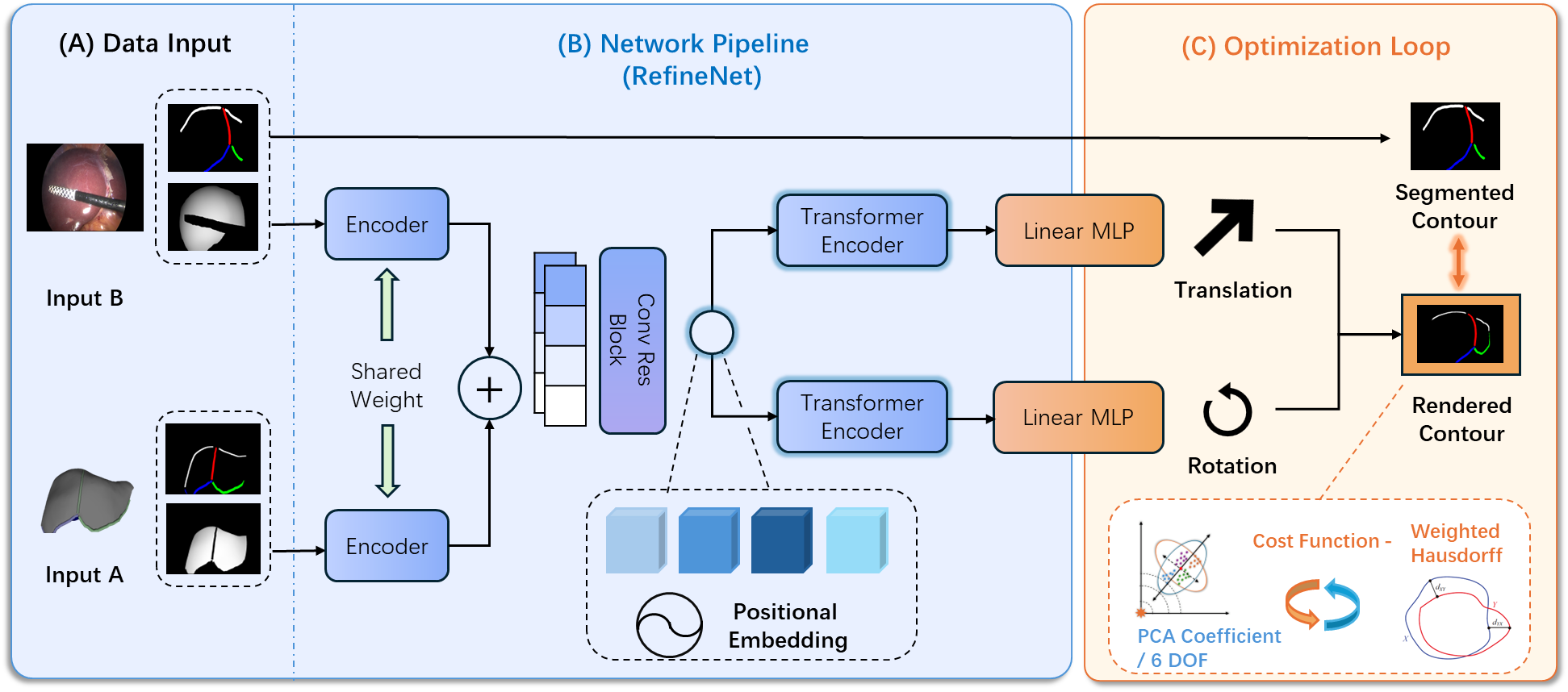}
    \caption{The registration workflow diagram: The first stage adopts the RefineNet component from FoundationPose by Wen et al., with the inputs replaced by the liver’s multi-class contour and depth images. The output is a relative pose update, which is used to initialize the subsequent non-rigid optimization.}
    \label{pipeline}
\end{figure}

\noindent\textbf{Statistical Deformation Modeling:}
We construct a patient-specific statistical deformation model using the liver-shape dataset of Montana-Brown et al.~\cite{montana2023saramis}. After rigidly aligning samples to the patient’s preoperative liver mesh using ICP, we register the patient mesh (source) to each aligned dataset mesh (target) with Non-rigid ICP, following the implementation of Foti et al.~\cite{foti2020intraoperative,foti2020meshpreprocessing}. This produces a set of deformed liver instances with the same topology as the patient mesh.

\begin{figure}[t]
    \centering
    \begin{minipage}[c]{0.63\linewidth}
        \centering
        \includegraphics[width=\linewidth]{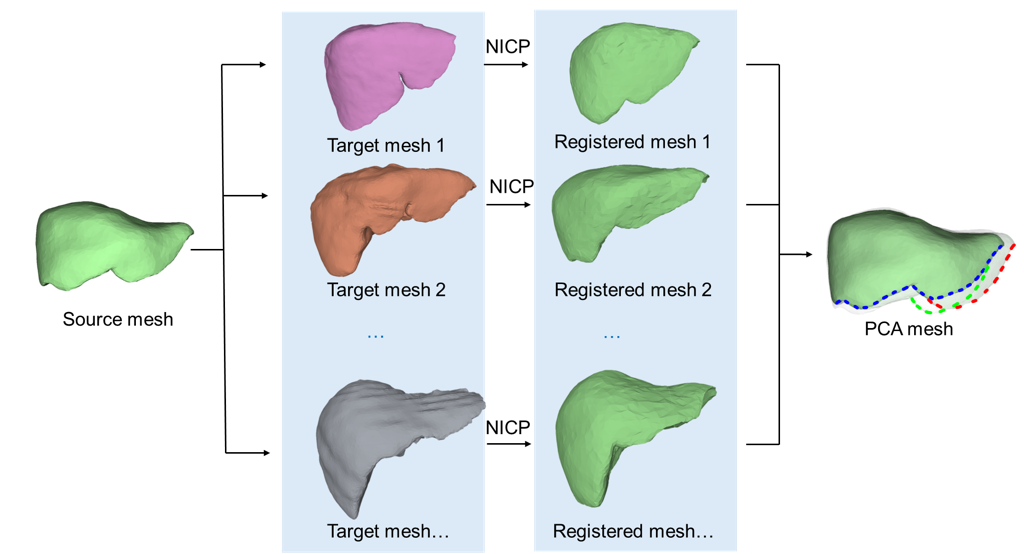}
    \end{minipage}\hfill
    \begin{minipage}[c]{0.35\linewidth}
        \scriptsize
        \textbf{NICP+PCA.}\\
        NICP establishes dense correspondence; PCA yields low-dimensional deformation modes.
    \end{minipage}
    \caption{Patient-specific statistical deformation model constructed from NICP-registered liver shapes.}
    \label{fig:NICP}
\end{figure}

We use the optimal-step NICP formulation of Amberg et al.~\cite{amberg2007optimal}, which estimates per-vertex affine transformations by balancing local rigidity and data fitting. At each iteration, correspondences are obtained by nearest-point search and filtered by normal consistency, i.e., $w_i=\gamma$ if $\mathbf{n}_i \cdot \mathbf{n}_i^{*}>\theta$ (with $\theta=0.7$), and $w_i=0$ otherwise. A coarse-to-fine rigidity schedule $\alpha \in [20,10,5,2,1,0.5,0.2]$ is used to progressively relax deformation constraints. The affine parameters are solved via regularized least squares, $\mathbf{x}^{(k+1)} = \arg\min_{\mathbf{x}} \|\mathbf{A}\mathbf{x}-\mathbf{b}\|_2^2 + \lambda\|\mathbf{x}\|_2^2$ with $\lambda=10^{-6}$, where $\mathbf{x}$ stacks per-vertex affine parameters and $(\mathbf{A},\mathbf{b})$ are assembled from rigidity and correspondence terms. Updated vertex positions are computed from the solved affine parameters.

For each registered instance, we compute per-vertex displacements relative to the canonical mesh and stack them into a data matrix. PCA on these displacements yields a low-dimensional deformation subspace. Let \(\mathbf{x}_0\in\mathbb{R}^{3V}\) denote the canonical shape (stacked vertex coordinates). With PCA components \(\mathbf{U}\) and standard deviations \(\boldsymbol{\sigma}\), the deformed shape is parameterized as \(\mathbf{x}(\boldsymbol{\beta})=\mathbf{x}_0+\mathbf{U}\,\mathrm{diag}(\boldsymbol{\sigma})\,\boldsymbol{\beta}\), where the normalized coefficients are clamped to \(\boldsymbol{\beta}\in[-1,1]^K\) to avoid unrealistic extrapolation. We retain the first ten principal modes ($K=10$) to define the deformation space. 
These modes explain approximately $75\%$ of the shape variance while keeping the deformation space compact and optimization stable.

\noindent\textbf{Joint Pose--Shape Optimization:}
We jointly optimize a 6-DoF rigid pose and the PCA deformation coefficients, i.e., the parameter vector comprises 3D translation, 3D rotation, and \(\boldsymbol{\beta}\). We refine these parameters by minimizing a weighted multi-class Hausdorff distance between rendered model contours and input contours. Let $n\in\{\text{ridge}_R,\text{ridge}_L,\text{lig},\text{sil}\}$ denote contour channels, and let $L_n$ be the set of pixels labeled as class $n$ in the 2D intraoperative annotation. To balance contributions across structures, we define channel weights as
$
w_n=\frac{|L_n|}{\sum_m |L_m|}
$
if $\sum_m |L_m|>0$, and $w_n=\frac{1}{4}$ otherwise. The objective is
\begin{equation}
\mathcal{L}(\mathbf{T},\boldsymbol{\beta})
=\sum_n w_n\,
d_H\!\Big(\mathcal{C}_n(\mathbf{T},\boldsymbol{\beta}),\,\mathcal{C}_n^{\mathrm{gt}}\Big),
\end{equation}


where $\boldsymbol{\beta}\in\mathbb{R}^{10}$ denotes the PCA shape coefficients, $\mathcal{C}_n(\cdot)$ is the rendered contour of class $n$, and $\mathcal{C}_n^{\mathrm{gt}}$ is extracted from $L_n$. We employ $d_H$ because it is sensitive to anatomical boundary alignment~\cite{zhang2025deep}. Although the resulting objective is non-smooth, in practice the proposed rigid initialization provides a suitable starting point for local refinement. We therefore optimize the box-constrained objective using L-BFGS-B with finite-difference gradients, together with bounded updates and a coarse-to-fine schedule for improved stability. Specifically, we first optimize the rigid pose while fixing $\boldsymbol{\beta}$, and then jointly refine pose and shape.

\section{Experiments and Results}

\noindent\textbf{Datasets and Implementation Details:}
For each patient, we define a reference pose for synthetic data generation such that at least three contour classes are visible. We then generate a synthetic training set by sampling \(5\times 10^5\) perturbed poses around this reference, with translations uniformly drawn from \([-50,50]\)~mm and rotations from \([-20^{\circ},20^{\circ}]\). For each pose, we render multi-class contours and depth (via the z-buffer). Samples with fewer than two visible contour types are discarded.

To reduce the synthetic-to-real gap, we apply modality-specific augmentations to both contour and depth inputs. These augmentations simulate fragmented contours, occlusions, mild elastic deformation, and noisy or missing depth.

\noindent\textbf{Data Preparation and Training Setup:} For the construction of the non-rigid deformation model, after removing incomplete liver meshes from the dataset of Montana-Brown et al.~\cite{montana2023saramis}, we retain 398 liver meshes from distinct subjects to extract the PCA basis.
For the RefineNet training, the remaining synthetic data are split into 90\%/10\% for training/validation. We train with Adam using a learning rate of \(1\times 10^{-4}\), batch size 32, for 50 epochs. The checkpoint with the lowest validation loss is selected. Training is conducted on NVIDIA GH200 GPUs, and inference is performed on an RTX 4070 Laptop GPU with an Intel i9-14900HX CPU.

\noindent\textbf{Experimental Setup:}
We evaluate our pipeline on 8 clinical cases in total. For quantitative analysis, we use the 4 cases from Rabbani et al.~\cite{rabbani2022methodology}, a public clinical benchmark that provides intraoperative ultrasound-localized tumors for target registration error (TRE) calculation. Although quantitatively annotated clinical datasets for tumor-level TRE remain scarce, this benchmark enables controlled comparison with prior work. We further include 4 external clinical cases for qualitative assessment of cross-dataset generalization. To rigorously isolate registration mechanics from downstream segmentation noise, we utilize manual contours as a controlled baseline (aligning with \cite{labrunie2022automatic,mhiri2025neural}). Depth is estimated via Depth Anything V2. This controlled setting isolates the registration component, but does not capture error propagation from automatic segmentation~\cite{lin2025subsampled,xu2025lapfm}.

For each patient, we perform 10 runs with different random initializations. In each run, TRE is first averaged over all frames of that patient using the public evaluation protocol, yielding one case-level mean TRE. We then report the mean and standard deviation over the 10 case-level mean TRE values to quantify robustness to initialization. Following prior studies, Patient~2 is known to be particularly challenging due to substantial intraoperative deformation/torsion and limited visible surface, which leads to consistently larger errors across methods~\cite{adagolodjo2017silhouette,labrunie2022automatic,labrunie2023automatic,kalantari2025stronger}. To enable fair comparison with existing literature, we report results both on all four patients and with Patient~2 excluded (w/o P2).
\begin{table}[t]
  \centering
  \caption{Registration performance in TRE (mm, lower is better). Best results are in \textbf{bold}. MA: manual initialization. For each patient, TRE is first averaged over all available frames using the public benchmark evaluation code. For the proposed method, we additionally report mean$\pm$std over 10 runs with different random initializations, where each run yields one case-level mean TRE.}
  \label{tab:tre_results}
  \resizebox{\textwidth}{!}{%
  \scriptsize
  \setlength{\tabcolsep}{4pt}
  \begin{tabular}{l c c c c c c c}
    \toprule
    \textbf{Method} & \textbf{Annotation} & \textbf{P1} & \textbf{P2} & \textbf{P3} & \textbf{P4} & \textbf{Avg} & \textbf{Avg w/o P2} \\
    \midrule
    MA & N/A & 15.14 & 35.48 & 30.48 & 16.29 & 24.35 & 20.63 \\
    Sil-B~\cite{adagolodjo2017silhouette} & Manual & \textbf{8.25} & 37.25 & 28.40 & 15.83 & 22.43 & 17.47 \\
    LMR~\cite{labrunie2023automatic} & Manual & 17.40 & 53.80 & 17.60 & 17.00 & 26.45 & 17.33 \\
    NM~\cite{mhiri2025neural} & Manual & 14.82 & 51.43 & 20.15 & 12.95 & 24.83 & 15.87 \\
    Opt-B~\cite{labrunie2022automatic} & Manual & 14.87 & N/A & 22.40 & \textbf{7.23} & N/A & 14.83 \\
    ADeLiR~\cite{gadoux2025automatic} & Auto & 12.45 & 46.44 & 17.13 & 11.20 & 21.81 & 13.60 \\
    ADeLiR~\cite{gadoux2025automatic} & Manual & 9.62 & 42.90 & 26.55 & 28.15 & 26.81 & 21.44 \\
    Proposed(CMA-ES)& Manual & \textbf{8.25$\pm$3.24} & {31.71$\pm$2.49} & {8.90$\pm$1.01} & 11.89$\pm$0.26 & {15.19} & {9.48} \\
    Proposed(L-BFGS-B) & Manual & {9.59$\pm$3.25} & \textbf{31.49$\pm$2.91} & \textbf{7.76$\pm$2.48} & 10.08$\pm$1.59 & \textbf{14.73} & \textbf{9.14} \\
    \bottomrule
  \end{tabular}
   }
\end{table}

\noindent\textbf{Main Results:}
Because prior methods use different annotation settings (manual versus automatic), Table~\ref{tab:tre_results} should be interpreted primarily as a benchmark-oriented reference rather than a strictly matched comparison.
The quantitative results are summarized in Table~\ref{tab:tre_results}.
Under the controlled manual-contour setting, the proposed method achieves the lowest average TRE among the methods evaluated with manual contour input.
Notably, Patient~2 has been consistently challenging in previous studies, where automatic methods yield substantially larger errors than manual alignment. In contrast, our method reduces the error on Patient~2 and achieves lower TRE than manual initialization on this challenging case. For comparison, we also evaluate CMA--ES under the same objective, initialization, and parameter bounds in Table~\ref{tab:tre_results}.

\begin{figure}[t]
    \centering
    \includegraphics[width=1\linewidth]{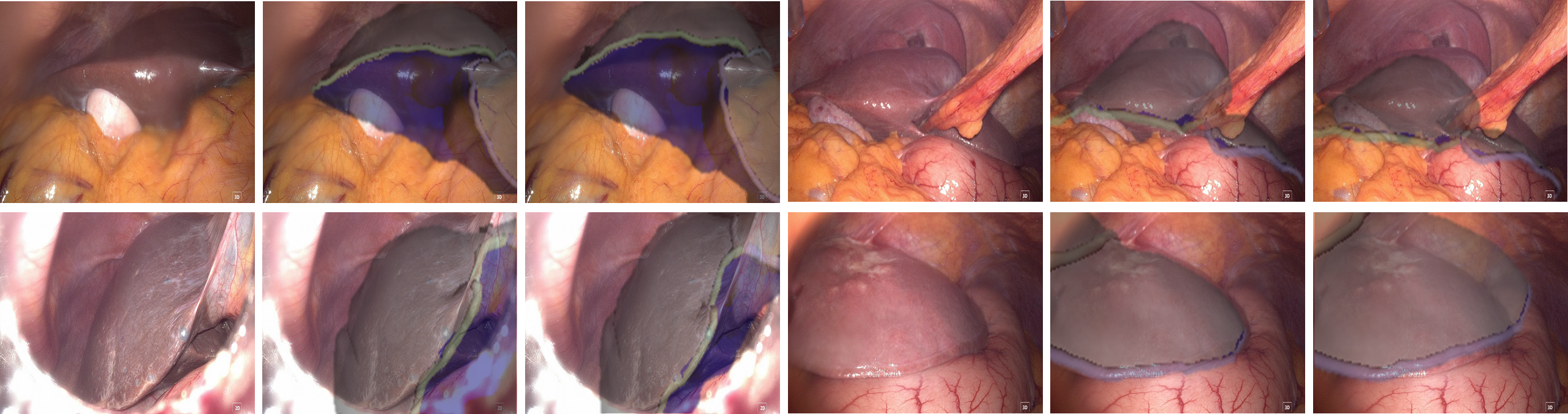}
    \caption{Qualitative overlays on four cases from an independent laparoscopic liver dataset, comparing CMA--ES and L-BFGS-B optimization. In each group, the leftmost column shows the original intraoperative image, followed by CMA--ES and L-BFGS-B results.}
    \label{fig:visualization}
\end{figure}

To complement the small quantitative benchmark, we additionally test the proposed pipeline on four cases from an independent laparoscopic liver dataset for qualitative cross-dataset assessment. Since tumor-level ground truth is unavailable, we provide qualitative overlays only (Fig.~\ref{fig:visualization}). The predicted registrations are visually plausible and broadly consistent with manual alignment in terms of surface contour consistency and overall pose, suggesting potential utility as an initialization aid in practice.

\noindent\textbf{Ablation Studies:}
We perform two ablations (Table~\ref{tab:ablation_all}). First, we compare three input variants for the rigid initialization network—contours only (C), contours+mask (C+M), and contours+masked depth (C+D)—under identical settings, and report the final end-to-end TRE after the same downstream refinement. The C+M variant is included as a control for the masking effect. Both \textbf{C+M} and \textbf{C+D} improve over \textbf{C}, and \textbf{C+D} achieves the best overall mean TRE and improves performance on 3/4 patients, indicating that depth provides useful geometric cues for resolving pose ambiguity.

\begin{table}[t]
  \centering
  \caption{Ablation studies (TRE in mm, lower is better). Best in \textbf{bold}.}
  \label{tab:ablation_all}
  \scriptsize
  \setlength{\tabcolsep}{5pt}
  \begin{tabular}{l c c c c c}
    \toprule
    \textbf{Method} & \textbf{P1} & \textbf{P2} & \textbf{P3} & \textbf{P4} & \textbf{Mean} \\
    \midrule
    \multicolumn{6}{l}{\textbf{(a) Input modality}} \\
    \midrule
    Contour & 15.25 & 39.48 & 18.94 & 20.21 & 23.47 \\
    Contour+Mask & 14.59 & \textbf{28.00} & 12.52 & 31.12 & 21.56 \\
    Contour+Depth & \textbf{13.68} & 32.32 & \textbf{11.03} & \textbf{18.18} & \textbf{18.80} \\
    \midrule
    \multicolumn{6}{l}{\textbf{(b) Tumor mapping}} \\
    \midrule
    FFD & 9.70 & 31.54 & \textbf{7.65} & 10.08 & 14.74 \\
    TPS & 9.75 & 31.51 & 7.74 & 10.08 & 14.77 \\
    Barycentric & \textbf{9.60} & \textbf{31.50} & 7.76 & 10.08 & \textbf{14.73} \\
    Linear & 9.74 & 31.51 & 7.68 & \textbf{10.07} & 14.75 \\
    RBF & 9.75 & \textbf{31.50} & 7.72 & 10.08 & 14.76 \\
    \bottomrule
  \end{tabular}
\end{table}

Second, we compare five tumor mapping strategies after non-rigid registration (FFD, TPS, Barycentric, Linear, and RBF). While the same surface alignment is used, tumor TRE varies across mappings, showing that target localization depends on deformation transfer. Barycentric mapping yields the lowest mean TRE (14.73\,mm), although differences across mappings are minimal compared to the size of the TRE.

\section{Discussion and Conclusion}

We presented a hybrid registration framework for AR-guided laparoscopic liver surgery that combines depth-augmented FoundationPose initialization with FE-free non-rigid refinement. The results show that monocular depth, even without accurate absolute scale, provides useful geometric cues for improving rigid pose estimation under limited visibility.

Our experiments also highlight an important system-level observation: improved surface alignment does not necessarily translate into lower tumor TRE. Even when the registered surface is similar, internal target localization can vary across deformation mappings, suggesting that clinically relevant AR accuracy depends not only on surface registration quality, but also on deformation-consistent target transfer.

This observation also supports our use of a compact deformation space with $K=10$ modes. Since the final tumor localization accuracy is partly limited by the propagation step, increasing model flexibility to achieve marginal gains in surface fitting may not yield proportional improvements in TRE, while increasing optimization cost and instability.

While the use of manual contours enables controlled evaluation of registration performance and facilitates comparison with prior baselines in Table~\ref{tab:tre_results}, it does not represent a fully automatic clinical pipeline. Our current results therefore isolate the registration component rather than the full end-to-end AR workflow.

In addition, L-BFGS-B achieved comparable or better accuracy than CMA-ES ($14.73$ vs.\ $15.19$~mm) while reducing runtime from approximately $30$--$60$~s to $2$--$3$~s per frame, yielding a substantially more favorable accuracy--runtime trade-off. This makes patient-specific non-rigid refinement more practical for intraoperative use. In navigation-enabled operating rooms with optical tracking, the camera--liver transformation can be updated after first-frame registration using tracked camera motion, while non-rigid refinement may be performed intermittently under the assumption of gradual deformation evolution~\cite{ramalhinho2025assessing}.

Future work will integrate automatic segmentation to evaluate end-to-end robustness under realistic noise, and will further investigate deformation-consistent tumor propagation for more reliable internal target localization. The implementation will be made publicly available upon acceptance.


%
%
%
%
\bibliographystyle{splncs04}
\bibliography{refs}




\end{document}